\documentclass[11pt,a4paper]{article}
\usepackage[hyperref]{acl2020}
\usepackage{times}
\usepackage{latexsym}

\usepackage{microtype}

\usepackage{graphicx}

\usepackage{multirow}
\usepackage{multicol}
\usepackage{amsmath}
\usepackage{amsfonts}
\usepackage{url}

\aclfinalcopy 
\def\aclpaperid{439} 

\title{Improving Entity Linking through Semantic Reinforced \\ Entity Embeddings}

\author{Feng Hou$^{\sharp\dagger}$, Ruili Wang$^{\sharp\dagger}$\thanks{\hspace{1mm}Corresponding author}, Jun He$^{\ddagger}$, Yi Zhou$^{\natural\dagger}$ \\
$^{\sharp}$\normalsize{School of Computer and Information Engineering, Zhejiang Gongshang University, Hangzhou, China}\\
  $^{\dagger}$School of Natural and Computational Sciences, Massey University, New Zealand \\
  $^{\ddagger}$School of Information Communication, National University of Defense Technology, China\\
  $^{\natural}$\normalsize{Shanghai Research Center for Brain Science and Brain-Inspired Intelligence, Zhangjiang Lab, China}\\
  \texttt{\{f.hou, ruili.wang\}@massey.ac.nz, prof.ruili.wang@gmail.com}\\
  \texttt{hejun\_nudt@nudt.edu.cn, yzhou@bsbii.cn }
  }
\date{}

\begin{document}
\maketitle
\begin{abstract}
  Entity embeddings, which represent different aspects of each entity with a single vector like word embeddings, are a key component of neural entity linking models. Existing entity embeddings are learned from canonical Wikipedia articles and local contexts surrounding target entities. Such entity embeddings are effective, but too distinctive for linking models to learn contextual commonality. We propose a simple yet effective method, FGS2EE, to inject fine-grained semantic information into entity embeddings to reduce the distinctiveness and facilitate the learning of contextual commonality. FGS2EE first uses the embeddings of semantic type words to generate semantic embeddings, and then combines them with existing entity embeddings through linear aggregation. Extensive experiments show the effectiveness of such embeddings. Based on our entity embeddings, we achieved new state-of-the-art performance on entity linking. 
\end{abstract}

\section{Introduction}

\begin{figure}[!h]
\centering
\includegraphics[width=0.4\textwidth]{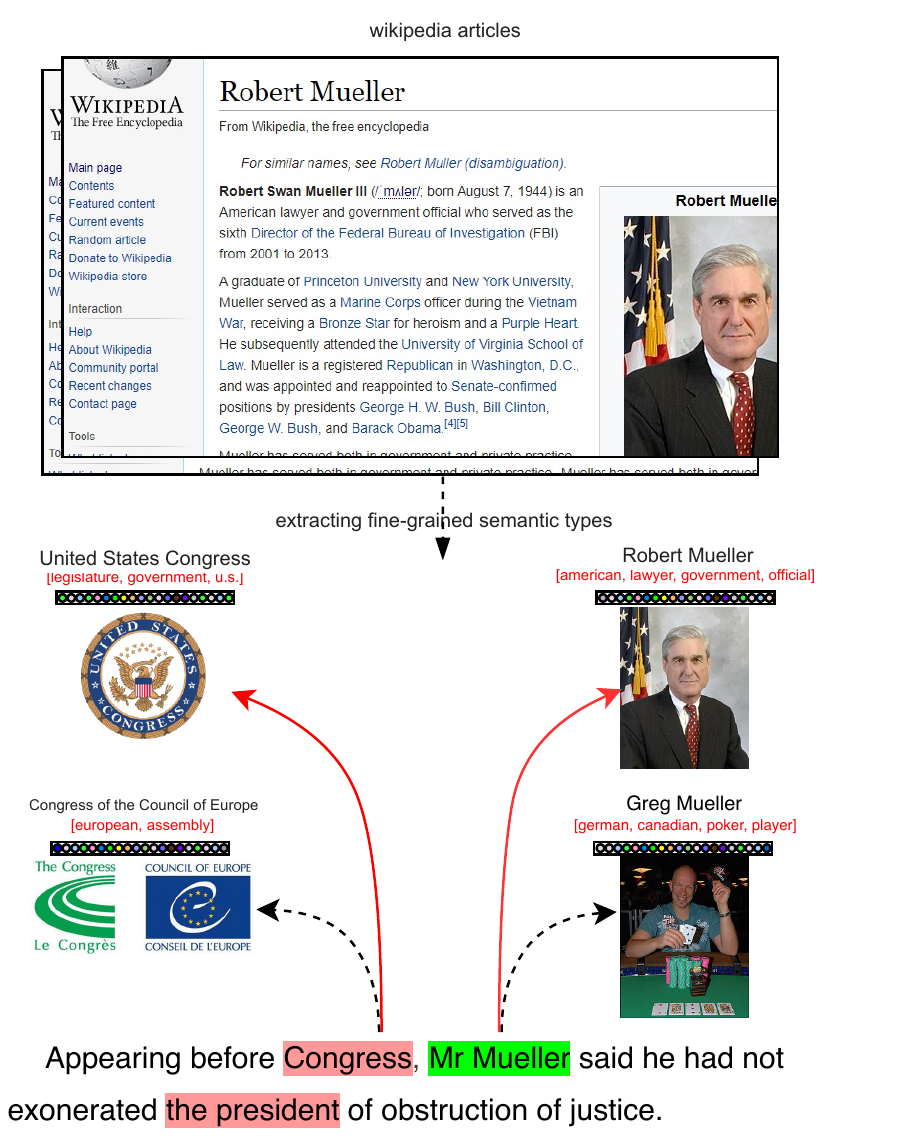}
\caption{Entity linking with embedded fine-grained semantic types}
\label{fig:el}
\end{figure}

Entity Linking (EL) or Named Entity Disambiguation (NED) is to automatically resolve the ambiguity of entity mentions in natural language by linking them to concrete entities in a Knowledge Base (KB). For example, in Figure \ref{fig:el}, mentions ``Congress'' and ``Mr. Mueller'' are linked to the corresponding Wikipedia entries, respectively.

Neural entity linking models use local and global scores to rank and select a set of entities for mentions in a document. Entity embeddings are critical for the local and global score functions. But the current entity embeddings \citep{ganea_localattention_2017} encoded too many details of entities, thus are too distinctive for linking models to learn contextual commonality. 

We hypothesize that fine-grained semantic types of entities can let the linking models learn contextual commonality about semantic relatedness. For example, \textit{rugby} related documents would have entities of \textit{rugby player} and \textit{rugby team}. If a linking model learns the contextual commonality of \textit{rugby} related entities, it can correctly select entities of similar types using the similar contextual information. 

In this paper, we propose a method FGS2EE to inject fine-grained semantic information into entity embeddings to reduce the distinctiveness and facilitate the learning of contextual commonality. FGS2EE uses the word embeddings of semantic words that represent the hallmarks of entities (e.g., \textit{writer, carmaker}) to generate semantic embeddings. We find that the training converges faster when using semantic reinforced entity embeddings. 

Our proposed FGS2EE consists of four steps: (i) creating a dictionary of fine-grained semantic words; (ii) extracting semantic type words from each entity’s Wikipedia article; (iii) generating semantic embedding for each entity; (iv) combining semantic embeddings with existing embeddings through linear aggregation.

\section{Background and Related Work}

\subsection{Local and Global Score for Entity Linking}

The local score $\Psi (e_i,c_i)$ \citep{ganea_localattention_2017} measures the relevance of entity candidates of each mention independently.
\begin{equation*}
    \Psi (e_i,c_i) = \mathbf{e}^{\top}_i \mathbf{B} \ f(c_i)
\end{equation*}
where $\mathbf{e}_i \in \mathbb{R}^d$ is the embedding of candidate entity $e_i$; $\mathbf{B} \in \mathbb{R}^{d \times d}$ is a diagonal matrix; $f(c_i) \in \mathbb{R}^d$ is a feature representation of local context $c_i$ surrounding mention $m_i$.

In addition to the local score, the global score adds a pairwise score $\Phi(e_i,e_j,D)$ to take the coherence of entities in document $D$ into account.
\begin{equation*}
\label{eq:pairwise}
    \Phi (e_i, e_j, D) = \frac{1}{n-1} \mathbf{e}_i^{\top} \mathbf{C}\ \mathbf{e}_j
\end{equation*}
where $\mathbf{e}_i$ and $\mathbf{e}_j \in \mathbb{R}^d$ are the embeddings of entities $e_i$, $e_j$, which are candidates for mention $m_i$ and $m_j$ respectively; $\mathbf{C} \in \mathbb{R}^{d \times d}$ is a diagonal matrix. The pairwise score of \citep{le_improvingel_2018} considers $K$ relations between entities.

\begin{equation*}
\label{eq:relations}
    \Phi (e_i, e_j, D) = \sum\limits_{k=1}^K \alpha_{ijk}\ \mathbf{e}_i^{\top} \mathbf{R}_k\ \mathbf{e}_j
\end{equation*}
where $\alpha_{ijk}$ is the weight for relation $k$, and $\mathbf{R}_k$ is a diagonal matrix for measuring relations $k$ between two entities.  

\subsection{Related Work}

Our research focuses on improving the vector representations of entities through fine-grained semantic types. Related topics are as follows.

\textbf{Entity Embeddings} Similar to word embeddings, entity embeddings are the vector representations of entities. The methods of
\citet{yamada-etal-2016-joint}, \citet{fang_joint_embedding_2016}, \citet{zwicklbauer_robust_2016}, use data about entity-entity co-occurrences to learn entity embeddings and often suffer from sparsity of co-occurrence statistics. \citet{ganea_localattention_2017} learned entity embeddings using words from canonical Wikipedia articles and local context surrounding anchor links. They used Word2Vec vectors \citep{tomas_mikolov_distributed_2013} of positive words and random negative words as input to the learning objective. Thus their entity embeddings are aligned with the Word2Vec word embeddings.

\textbf{Fine-grained Entity Typing} Fine-grained entity typing is a task of classifying entities into fine-grained types \citep{xiao_ling_fine-grained_2012} or ultra fine-grained semantic labels \citep{eunsol_choi_ultra-fine_2018}. \citet{rajarshi_bhowmik_generating_2018} used a memory-based network to generate a short description of an entity, e.g. ``Roger Federer'' is described as `Swiss tennis player'. In this paper, we heuristically extract fine-grained semantic types from the first sentence of Wikipedia articles. 

\textbf{Embeddings Aggregation} Our research is closely related to the work on aggregation and evaluation of the information content of embeddings from different sources (e.g., polysemous words have multiple sense embeddings), and fusion of multiple data sources \cite{wang2018review}. \citet{arora_linear_2018} hypothesizes that the global word embedding is a linear combination of its sense embeddings. They showed that senses can be recovered through sparse coding. \citet{mu_geometry_2016} showed that senses and word embeddings are linearly related and sense sub-spaces tend to intersect over a line. \citet{yaghoobzadeh_semantic_class_2019} probe the aggregated word embeddings of polysemous words for semantic classes. They created a WIKI-PSE corpus, where word and semantic class pairs are annotated using Wikipedia anchor links, e.g., ``apple'' has two semantic classes: \textit{food} and \textit{organization}. A separate embedding for each semantic class was learned based on the WIKI-PSE corpus. They found that the linearly aggregated embeddings of polysemous words represent well their semantic classes.

The most similar work is that of \citet{nitish_gupta_entity_2017}, but there are many differences: (i) they use the FIGER \citep{xiao_ling_fine-grained_2012} type taxonomy that contains manually curated 112 types organized into 2 levels; we employ over 3000 vocabulary words as type, and we treat them as a flat list; 
(ii) they mapped the Freebase types to FIGER types,but this method is less credible, as noted by \citet{daniel_gillick_contextdependent_2014}; we extract type words directly from Wikipedia articles, which is more reliable.
(iii) their entity vectors and type vectors are learned jointly on a limited corpus. Ours are linear aggregations of existing entity vectors, and word vectors learned from a large corpus, such fine-grained semantic word embeddings are helpful for capturing informative context. 

\subsection{Motivation}

Coarse-grained semantic types (e.g. \textit{person}) have been used for candidate selection \citep{ganea_localattention_2017}. We observe that fine-grained semantic words appear frequently as apposition (e.g., \textit{Defense contractor} Raytheon), coreference (e.g., the \textit{company}) or anonymous mentions (e.g., \textit{American defense firms}). These fine-grained types of entities can help capture local contexts and relations of entities. 

Some of these semantic words have been used for learning entity embeddings, but they are diluted by other unimportant or noisy words. We reinforce entity embeddings with such fine-grained semantic types.

\section{Extracting Fine-grained Semantic Types}

We first create a dictionary of fine-grained semantic types, then we extract fine-grained types for each entity.

\subsection{Semantic Type Dictionary}

We select those words that can encode the hallmarks of individual entities. Desiderata are as follows: 
\begin{itemize}
\itemsep-0.4em 
    \item profession/subject, e.g., \textit{footballer}, \textit{soprano, biology, rugby}.
    
    \item title, e.g., \textit{president, ceo, head, director}.
    
    \item industry/genre, e.g., \textit{carmaker}, \textit{manufacturer, defense contractor}, \textit{hip hop}.
    
    \item geospatial, e.g., \textit{canada, asian}, \textit{australian}.
    
    \item ideology/religion, e.g., \textit{communism, buddhism}.
    
    \item miscellaneous, e.g., \textit{book, film, tv, ship, language}.
    
\end{itemize}



We extract noun frequency from the first sentence of each entity in the Wikipedia dump. Then some seed words are manually selected from frequent nouns. We use word similarity to extend these seed words and finally got a dictionary with 3,227 fine-grained semantic words.

Specifically, we use \texttt{spaCy} to compute the similarity between words. For each seed word, we find the top 100 similar words that also appear in Wikipedia articles. We then manually select semantic words from these extended words.

\subsection{Extracting Semantic Types}

For each entity, we extract at most 11 dictionary words (phrases) from its Wikipedia article. For example, ``Robert Mueller'' in Figure \ref{fig:el} will be typed as [\textit{american, lawyer, government, official, director}].


\subsection{Remapping Semantic Words}

For some semantic words (e.g., \textit{conchologist}) or semantic phrases (e.g., \textit{rugby league}), there are no word embeddings available for generating the semantic entity embeddings. We remap these semantic words to semantically similar words that are more common. For example, the \textit{conchologist} is remapped to \textit{zoologist}, and \textit{rugby league} is remapped to \textit{rugby\_league}.

\section{FGS2EE: Injecting Fine-Grained Semantic Information into Entity Embeddings}

FGS2EE first uses semantic words of each entity to generate semantic entity embeddings, then combine them with existing entity embeddings to generate semantic reinforced entity embeddings.

\subsection{Semantic Entity Embeddings}

Based on the semantic words of each entity, we can produce a semantic entity embedding. We treat each semantic word as a sense of an entity. The embedding of each sense is represented by the Word2Vec embedding of the semantic word. Suppose we only consider $T$ semantic words for each entity, and the set of entity words of entity $e$ is denoted as $S_e$. Then the semantic entity embedding $\mathbf{e}^s$ of entity $e$ is generated as follows:
\begin{equation}
\label{eq:semantic}
    \mathbf{e}^s = \frac{1}{T}\sum\limits^{T}_{i=1} \mathbf{e}_{w_i}
\end{equation}
where $w_i \in S_e$ is the $i$th semantic word, $\mathbf{e}_{w_i}$ is the Word2Vec embedding\footnote{\url{https://code.google.com/archive/p/word2vec/}} of semantic word $w_i$. If $|S_e| < T$, then $T=|S_e|$.

\begin{table*}[t!]
\tiny
\centering
\begin{tabular}{ll|l|llllll}
  \textbf{Entity Embeddings} & \textbf{Linking Methods} & \textbf{AIDA-B} & \textbf{MSNBC} & \textbf{AQUAINT} & \textbf{ACE2004} & \textbf{CWEB} & \textbf{WIKI} & \textbf{Avg}\\
  \hline
  \multicolumn{2}{c|}{\textit{Wikipedia}} & & & & & & & \\
  -&\citep{milne_learning_2008} &- &78 &85 &81 &64.1 &81.7 & 77.96 \\
  -&\citep{ratinov_svm_2011} &-&75 &83&82&56.2&67.2 & 72.68 \\
  -&\citep{hoffart_robust_2011} &-&79 &56&80&58.6&63 & 67.32 \\
  -&\citep{cheng_ilp_2013} &-&90 &90 &86 &67.5 &73.4 &81.38 \\
  -&\citep{chisholm-hachey-2015-entity} &84.9&-&-&-&-&- & - \\
  \hline
  \multicolumn{2}{c|}{\textit{Wiki + Unlabelled documents}} & & & & & & & \\
  -&\citep{lazic_plato_2015} &86.4&-&-&-&-&- & - \\
  \citep{ganea_localattention_2017}&\citep{le-titov-2019-boosting} &\textbf{89.66$\pm$0.16} &92.2$\pm$0.2 &90.7$\pm$0.2 &\textbf{88.1$\pm$0.0} &78.2$\pm$0.2 &81.7$\pm$0.1 & 86.18 \\
 $T=6, \alpha = 0.1$ &\citep{le-titov-2019-boosting} &89.58$\pm$0.2 &\textbf{92.3$\pm$0.1} &90.93$\pm$0.2 &87.88$\pm$0.17 &\textbf{78.47$\pm$0.11} &81.71$\pm$0.21 & 86.26 \\
 $T=11, \alpha=0.2$ &\citep{le-titov-2019-boosting} &89.23$\pm$0.31 &92.15$\pm$0.24 &\textbf{91.22$\pm$0.18} &88.02$\pm$0.15 &78.29$\pm$0.17 &\textbf{81.92$\pm$0.36} & \textbf{86.32} \\
  \hline
  \multicolumn{2}{c|}{\textit{Wiki + Extra supervision}} & & & & & & &\\
  -&\citep{chisholm-hachey-2015-entity} &88.7&-&-&-&-&- & - \\
  \hline
  \multicolumn{2}{c|}{\textit{Fully-supervised(Wiki+ AIDA train)}} & & & & & & &\\
  -&\citep{guo_robust_2018} &89.0&92 &87 &88 &77 &\textbf{84.5} & 85.7 \\
  -&\citep{globerson_collective_2016} &91.0&- &- &- &- &- &- \\
  \citep{yamada-etal-2016-joint}&\citep{yamada-etal-2016-joint} &91.5&- &- &- &- &- &- \\
  \citep{ganea_localattention_2017}&\citep{ganea_localattention_2017} &92.22$\pm$0.14 &93.7$\pm$0.1 &\textbf{88.5$\pm$0.4} &88.5$\pm$0.3 &\textbf{77.9$\pm$0.1} &77.5$\pm$0.1 & 85.22 \\
  
  \citep{ganea_localattention_2017}&\citep{le_improvingel_2018} &93.07$\pm$0.27 &93.9$\pm$0.2 &88.3$\pm$0.6 &89.9$\pm$0.8 &77.5$\pm$0.1 &78.0$\pm$0.1 & 85.5 \\
  
  \citep{ganea_localattention_2017}& DCA \citep{yang-etal-2019-learning} &\textbf{93.73$\pm$0.2} &93.80$\pm$0.0 &88.25$\pm$0.4 &90.14$\pm$0.0 &75.59$\pm$0.3 &\textbf{78.84$\pm$0.2} & 85.32 \\
  
  $T=6, \alpha=0.1$ &\citep{le_improvingel_2018}  &92.29$\pm$0.21 &94.1$\pm$0.24 &88.0$\pm$0.38 &90.14$\pm$0.32 &77.23$\pm$0.18 &77.16$\pm$0.43 & 85.33 \\
  
  $T=11, \alpha = 0.2$ &\citep{le_improvingel_2018}  &92.63$\pm$0.14 &\textbf{94.26$\pm$0.17} &88.47$\pm$0.23 &\textbf{90.7$\pm$0.28} &77.41$\pm$0.21 &77.66$\pm$0.23 & \textbf{85.7} \\
  
\end{tabular}
\caption{F1 scores on six test sets. The last column is the average of F1 scores on the five out-domain test sets.}
\label{tab:results}
\end{table*}

\subsection{Semantic Reinforced Entity Embeddings}

We create a semantic reinforced embedding for each entity by linearly aggregating the semantic entity embeddings and Word2Vec style entity embeddings \citep{ganea_localattention_2017} (hereafter referred to as ``Wikitext entity embeddings'').

Our semantic entity embeddings tend to be homogeneous. If we average them with the Wikitext embeddings, the aggregated embeddings would be homogeneous too. Thus the entity linking model would not be able to distinguish between those similar candidates. Our semantic reinforced entity embedding is a weighted sum of semantic entity embedding and Wikitext entity embedding, similar to \citep{yaghoobzadeh_semantic_class_2019}. We use a parameter $\alpha$ to control the weight of semantic entity embeddings. Thus the aggregated (semantic reinforced) entity embeddings achieve a trade-off between homogeneity and heterogeneity. 

\begin{equation}
\label{eq:sri}
    \mathbf{e}^a = (1-\alpha)\ \mathbf{e}^w + \alpha \ \mathbf{e}^s
\end{equation}
where $\mathbf{e}^w$ is the Wikitext entity embedding of entity $e$.

\section{Experiments}

\subsection{Datasets and Evaluation Metric}

We use the Wikipedia dump 20190401 to extract fine-grained semantic type dictionary and semantic types for entities. We use the Wikitext entity embeddings shared by \citet{le_improvingel_2018,le-titov-2019-boosting}. For entity linking corpora, we use the datesets shared by \citet{ganea_localattention_2017} and \citet{le_improvingel_2018,le-titov-2019-boosting}.

We use the standard micro F1-score as evaluation metric. Our data and source code are publicly available at \texttt{\small github} \footnote{\url{https://github.com/fhou80/EntEmb/}}.

\subsection{Experimental Settings}

The parameters $T$ in Equation (\ref{eq:semantic}) and $\alpha$ in Equation (\ref{eq:sri}) are critical for the effectiveness of our semantic reinforced entity embeddings. We got two sets of entity embeddings with two combinations of parameters: $T=6, \alpha=0.1$ and $T=11, \alpha=0.2$

To test the effectiveness of our semantic reinforced entity embeddings, we use the entity linking models \textbf{mulrel} \citep{le_improvingel_2018} (ment-norm $K=3$) and \textbf{wnel} \citep{le-titov-2019-boosting} that are publicly available. We do not optimize their entity linking code. We just replace the entity embeddings with our semantic reinforced entity embeddings. 

Similar to \citet{ganea_localattention_2017} and \citet{le_improvingel_2018,le-titov-2019-boosting}, we run our system 5 times for each combination of entity embeddings and linking model, and report the mean and 95\% confidence interval of the micro F1 score.

\subsection{Results}

\begin{figure}[!h]
\centering
\includegraphics[width=0.4\textwidth]{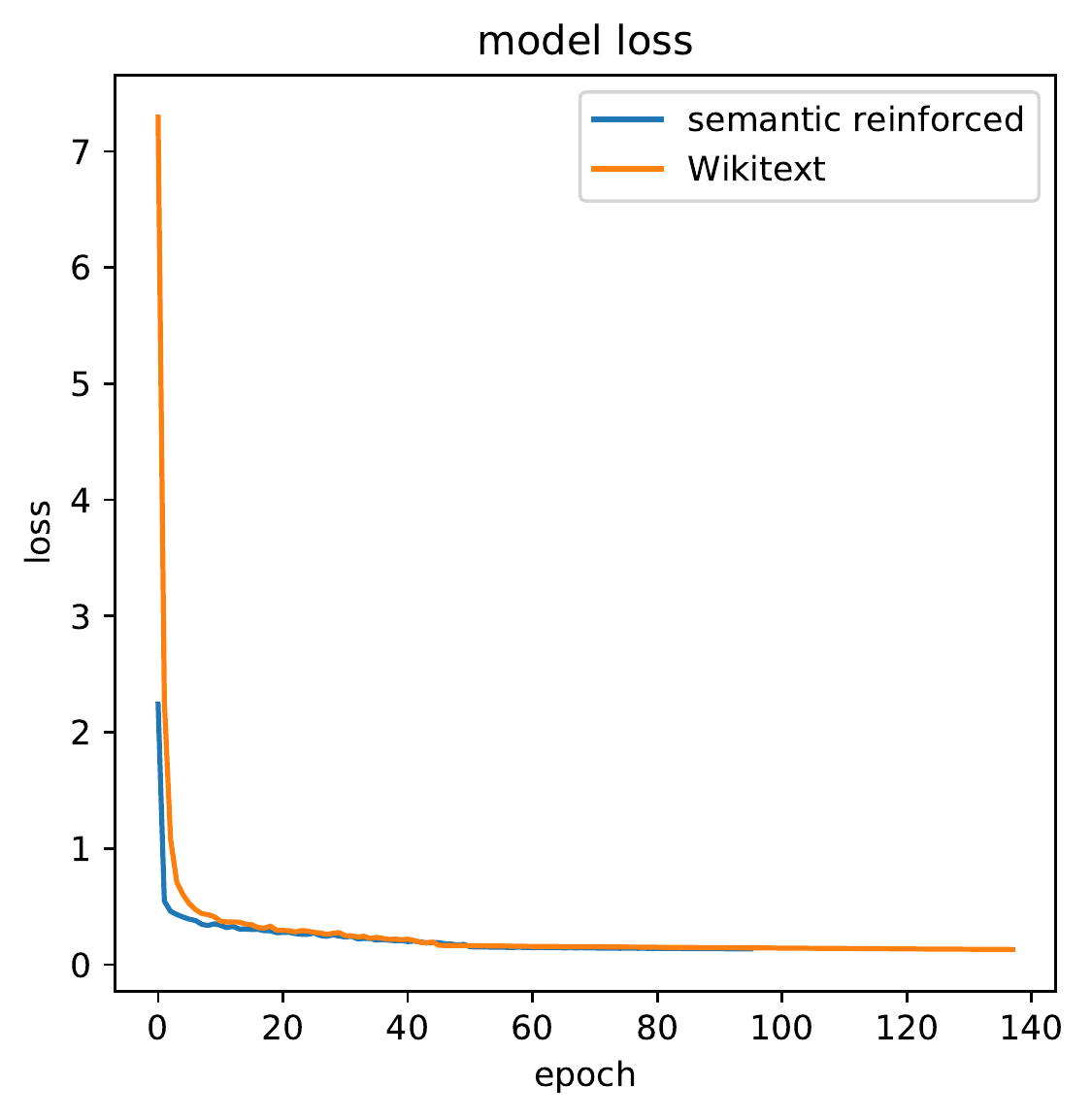}
\caption{Learning curves of \textbf{mulrel} \citep{le_improvingel_2018} using two different sets of entity embeddings.}
\label{fig:curve}
\end{figure}

\begin{figure}[!h]
\centering
\includegraphics[width=0.45\textwidth]{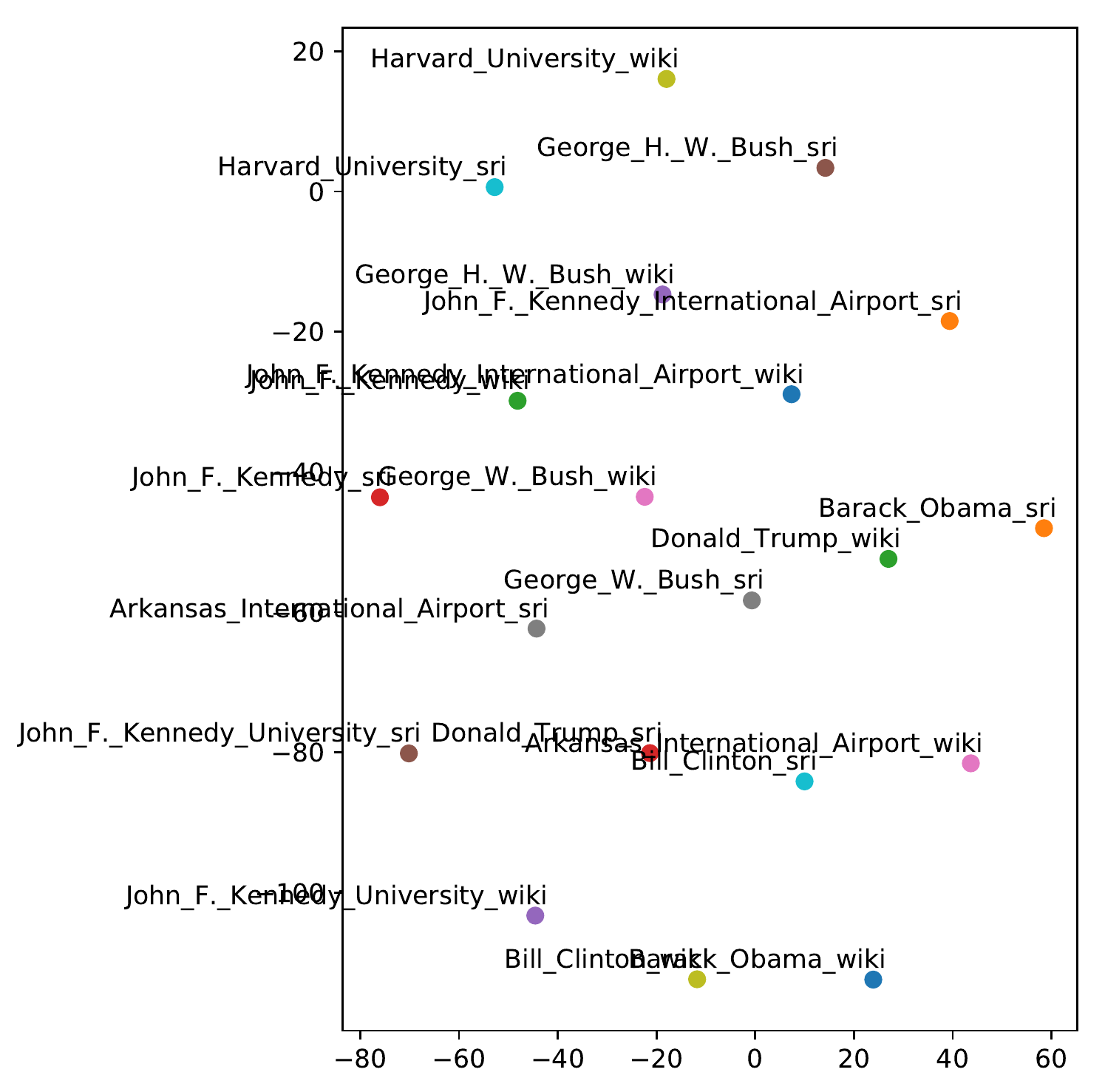}
\caption{T-SNE visualization of two sets of entity embeddings. Suffix ``\_wiki'' denotes the Wikitext entity embeddings, while suffix ``\_sri'' denotes the semantic reinforced entity embeddings ($T=11, \alpha=0.2$).}
\label{fig:tsne}
\end{figure}

The results on six testing datasets are shown in Table \ref{tab:results}. For the \textbf{mulrel} model, our entity embeddings ($T=11, \alpha=0.2$) improved performance drastically on MSNBC, ACE2004 and average of out-domain test sets. Be aware that CWEB and WIKI are believed to be less reliable \citep{ganea_localattention_2017}. For the \textbf{wnel} model, our both sets of entity embeddings are more effective for four of the five out-domain test sets and the average.

Our entity embeddings are better than that of \citet{ganea_localattention_2017} when tested on the \textbf{mulrel} \citep{le_improvingel_2018} (ment-norm $K=3$) and \textbf{wnel} \citep{le-titov-2019-boosting} entity linking models. \citet{ganea_localattention_2017} showed that their entity embeddings are better than that of \citet{yamada-etal-2016-joint} using the entity relatedness metrics.

One notable thing for our semantic reinforced entity embeddings is that the training using our entity embeddings converges much faster than that using Wikitext entity embeddings, as shown in Figure \ref{fig:curve}. One reasonable explanation is that the fine-grained semantic information lets the linking models capture the commonality of semantic relatedness between entities and contexts, hence facilitate the training.

The properties of two different sets of entity embeddings can be visually manifested in Figure \ref{fig:tsne}. Our semantic reinforced entity embeddings draw entities of similar types closer, and entities of different types further. For example, our semantic reinforced embeddings of ``John F. Kennedy University'' and ``Harvard University'' are closer than the Wikitext embeddings, while our embeddings of ``John F. Kennedy International Airport'' and ``John F. Kennedy'' are further. We believe this property contributes to the faster convergence.

\section{Conclusion}

In this paper, we presented a simple yet effective method, FGS2EE, to inject fine-grained semantic information into entity embeddings to reduce the distinctiveness and facilitate the learning of contextual commonality. FGS2EE first uses the word embeddings of semantic type words to generate semantic embeddings, and then combines them with existing entity embeddings through linear aggregation. Our entity embeddings draw entities of similar types closer, while entities of different types are drawn further. Thus can facilitate the learning of semantic commonalities about entity-context and entity-entity relations. We have achieved new state-of-the-art performance using our entity embeddings.

For the future work, we are planning to extract fine-grained semantic types from unlabelled documents and use the relatedness between the fine-grained types and contexts as distant supervision for entity linking.

\bibliography{acl2020}
\bibliographystyle{acl_natbib}

\end{document}